\documentclass[journal]{IEEEtran}

\ifCLASSINFOpdf
\else
 \usepackage[dvips]{graphicx}
\fi
\usepackage{url}
\usepackage{booktabs}

\usepackage{graphicx}
\usepackage{adjustbox}
\usepackage{makecell}
\usepackage{float}

\usepackage{caption}
\usepackage[dvipsnames]{xcolor}
\usepackage[square,numbers,sort&compress]{natbib}
\usepackage{amsmath}
\usepackage{amssymb}
\usepackage{bbm}

\usepackage[colorlinks=true]{hyperref}
\hypersetup{%
,urlcolor=blue
,citecolor=blue
,linkcolor=red
}
\setcitestyle{square}

\newcommand{\calX}{\mathcal{X}}
\newcommand{\calY}{\mathcal{Y}}

\newcommand{\calD}{\mathcal{D}}

\newcommand{\by}{\mathbf{y}}

\newcommand{\R}{\mathbb{R}}

\usepackage{amsthm}

\usepackage{tikz}
\usetikzlibrary{arrows}

\usepackage{multirow}
\usepackage[T1]{fontenc}

\usepackage{times}
\usepackage{epsfig}
\usepackage{graphicx}
\usepackage{amsmath}
\usepackage{amsthm}
\usepackage{amssymb}
\usepackage{multirow}
\usepackage{xcolor}
\usepackage{soul}
\usepackage{adjustbox}
\usepackage[numbers]{natbib}

\newcommand{\D}{\mathcal{D}}

\def\vh{{\boldsymbol{h}}}

\def\vp{{\boldsymbol{p}}}

\theoremstyle{definition}

\usepackage{graphicx}
\usepackage{comment}
\usepackage{epsfig}
\usepackage{graphicx}
\usepackage{multirow}
\usepackage{xcolor}
\usepackage{soul}
\usepackage{enumerate}
\usepackage{amsmath,amssymb} %
\usepackage{xspace}
\usepackage{cleveref}
\usepackage{subcaption}
\usepackage{wrapfig}
\usepackage{bm}

\renewcommand{\paragraph}[1]{\vspace{.5em} \noindent{\textbf{#1}}}

\makeatletter
\DeclareRobustCommand\onedot{\futurelet\@let@token\@onedot}
\def\@onedot{\ifx\@let@token.\else.\null\fi\xspace}

\makeatother

\definecolor{applegreen}{rgb}{0.55, 0.71, 0.0}
\definecolor{asparagus}{rgb}{0.53, 0.66, 0.42}
\definecolor{ao}{rgb}{0.0, 0.5, 0.0}

\newcommand{\ntkr}{{\bm{\Theta}(\calD_r, \calD_r)}}
\newcommand{\ntkf}{{\bm{\Theta}(\calD_f, \calD_f)}}
\newcommand{\ntkrf}{{\bm{\Theta}(\calD_r, \calD_f)}}

\begin{document}

\title{Fast-NTK: Parameter-Efficient Unlearning for Large-Scale Models}

\author{
Guihong Li, Hsiang Hsu, Chun-Fu Chen, and Radu Marculescu
\thanks{Guihong Li and Radu Marculescu are with the University of Texas at Austin, Austin, TX 78712 USA. This work was done during an internship at JPMorgan Chase \& Co. (e-mails: lgh@utexas.edu and radum@utexas.edu).}
\thanks{Hsiang Hsu and Chun-Fu Chen are with JPMorgan Chase \& Co., USA (e-mails: hsiang.hsu@jpmchase.com and richard.cf.chen@jpmchase.com).}
}

% \markboth{Journal of \LaTeX\ Class Files, Vol. 14, No. 8, August 2015}
% {Shell \MakeLowercase{\textit{et al.}}: Bare Demo of IEEEtran.cls for IEEE Journals}
\maketitle

\begin{abstract}
The rapid growth of machine learning has spurred legislative initiatives such as ``the Right to be Forgotten,'' allowing users to request data removal. 
In response, ``machine unlearning'' proposes the selective removal of unwanted data without the need for retraining from scratch. 
While the Neural-Tangent-Kernel-based (NTK-based) unlearning method excels in performance, it suffers from significant computational complexity, especially for large-scale models and datasets.
Our work introduces ``Fast-NTK,'' a novel NTK-based unlearning algorithm that significantly reduces the computational complexity by incorporating parameter-efficient fine-tuning methods, such as fine-tuning batch normalization layers in a CNN or visual prompts in a vision transformer. 
Our experimental results demonstrate scalability to much larger neural networks and datasets (e.g., 88M parameters; 5k images), surpassing the limitations of previous full-model NTK-based approaches designed for smaller cases (e.g., 8M parameters; 500 images). 
Notably, our approach maintains a performance comparable to the traditional method of retraining on the retain set alone. 
Fast-NTK can thus enable for practical and scalable NTK-based unlearning in deep neural networks.
\end{abstract}


\IEEEpeerreviewmaketitle

\section{Introduction}

\IEEEPARstart{T}{he} surge in machine learning applications has prompted legislative actions, notably ``the Right to be Forgotten,'' allowing individuals to request the removal of their online information \cite{eu_gdpr}.
However, the privacy challenge remains as erasing data from databases may persist in machine learning models, particularly in deep neural networks (DNNs), which are recognized for their efficient training data memorization \cite{iclr_gan_memory}.
To address this issue, ``machine unlearning'' has emerged to enable selective removal of unwanted ``forget samples'' without the need of retraining the model from scratch \citep{unlearn_survey}.

Among various unlearning algorithms \cite{sisa_unlearn, gupta2021adaptive, neel2021descent, tarun2023deep, grad_up, chen2019novel}, neural-tangent-kernel-based (NTK-based) unlearning stands out for its state-of-the-art performance~\cite{golatkar2020forgetting,golatkar2020eternal}. 
However, NTK-based unlearning algorithms are challenging due to the need of computing kernel matrices with respect to all samples and model weights. 
This computational complexity grows polynomially with the number of samples and model weights, thus resulting in intensive computation costs and memory consumption. 
Consequently, the effectiveness of NTK-based unlearning algorithms is often limited only to small-scale models and datasets (e.g., 8M parameters and 500 images). 

\begin{figure}
 \centering
 \includegraphics[width=0.4\textwidth]{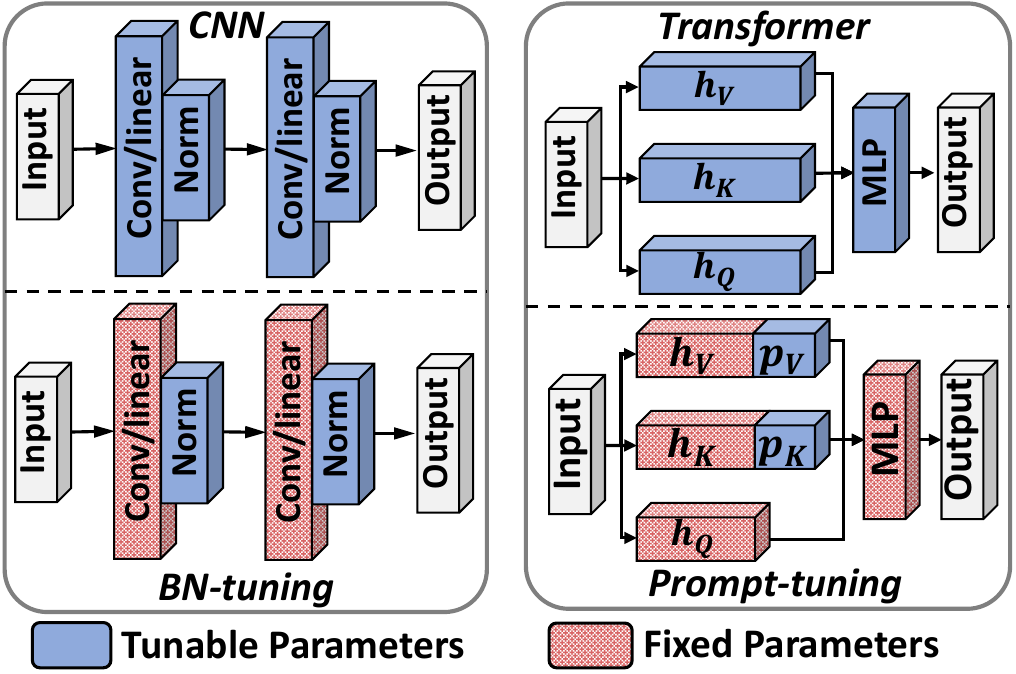}
 \caption{\footnotesize A schematic representation of parameter-efficient fine-tuning and unlearning. For CNNs (left), instead of updating the entire model, we conduct the fine-tuning and NTK-based unlearning on the batch normalization (BN) layers. For transformers (right), we only modify the prompts ($\vp_K$ and $\vp_V$) appended to the entire model.
 }\vspace{-5mm}
 \label{fig:overview}
\end{figure}

In this letter, we draw inspiration from recent strides in parameter-efficient fine-tuning (PEFT) \cite{lin2022device_256k,zheng2022_vit_prompt,jia2022visual_vit_prompt,chiang2023mobiletl} and leverage the NTK-based unlearning algorithms---specifically, the computation of kernel matrices---to work with a limited set of important parameters, such as those in batch normalization layers and visual prompts. We term this approach ``Fast-NTK,'' as shown in Figure~\ref{fig:overview}.
Unlike the conventional application of NTK-based unlearning algorithms across all model weights, Fast-NTK significantly reduces the parameter count (cf.~Table~\ref{tab:cnn_results}) of the standard implementation of the entire model. 
Remarkably, our experimental results, e.g., vision transformers (ViTs) on the ImageNet-R dataset, demonstrate indistinguishable performance compared to the commonly-used baseline that retrains the model from scratch only on the remaining data. 
Consequently, we believe our approach can enable a practical and scalable paradigm for the NTK-based unlearning approaches.\footnote{Codes to reproduce our experiments will be made public.}

\vspace{-1.5mm}
\section{Background and Related Work}\label{sec:background-related-work}

Consider a training dataset $\calD$ that can be divided into two disjoint subsets: a forget set $\calD_f$ which is the target for unlearning, and a retain set $\calD_r$ which collects the remaining samples.
The objective of machine unlearning is to eliminate the knowledge from the forget samples in $\calD_f$ of a model trained with $\calD$, while minimizing the performance degradation of the retain samples in $\calD_r$ \cite{mul_survey2023}.
One intuitive strategy is to retrain the entire model from scratch, utilizing only the samples in $\calD_r$.
However, this process can be time-consuming, particularly when dealing with large-scale datasets and models. 
Consequently, current research endeavors to directly erase the knowledge associated with the forget samples from the model, without necessitating a complete retraining.

There exist three distinct strategies for accomplishing machine unlearning: data partitioning~\cite[SISA]{sisa_unlearn}, mimicking differential privacy~\cite{gupta2021adaptive}, and adjusting the model weights~\cite{neel2021descent, tarun2023deep, grad_up, chen2019novel}.
This letter specifically delves into the intricacies of updating the model weights, hence targeting machine unlearning through the computation of Neural Tangent Kernels (NTKs) \cite{lee2019wide,jacot2018neural}.

Consider a neural network $f_\theta: \calX \to \calY$, parameterized by $\theta \in \R^d$, where $\calX$ and $\calY$ are the support sets of the input and output, respectively.
The NTK matrix of the two datasets $\calD_1$ and $\calD_2$ is defined as $\bm{\Theta}(\calD_1, \calD_2) \triangleq \nabla_{\theta} f_\theta(\calD_1) \nabla_{\theta} f_\theta(\calD_2)^\top$. 
Let $\theta$ and $\theta_{r}$ be the weights from training with the entire training set $\calD$ and the retain set $\calD_r$ alone, respectively.
Note that directly obtaining $\theta_{r}$ from $\theta$ is the goal of machine unlearning by updating model weights. 
By linearizing the outputs of $f_\theta$, we can approximate $\theta$ and $\theta_{r}$ in closed forms, and directly move the model weights from $\theta$ to $\theta_r$ by an optimal one-shot update:  
\vspace{-1mm}
\begin{equation}\label{eq:optimal-scrubbing}
\theta_r=\theta + \bm{P} \nabla_{\theta} f_\theta(\calD_f)^\top \bm{M} \bm{V},
\end{equation}
where $\bm{P} = \bm{I} - \nabla_{\theta} f_\theta(\calD_r)^\top {\bm{\Theta}(\calD_r, \calD_r)}^{-1}\nabla_{\theta} f_\theta(\calD_r)$ is the matrix that projects the gradients of the samples to forget $\nabla_{\theta} f_\theta(\calD_f)$ to a space that is orthogonal to the space spanned by the gradients of all retain samples; $\bm{M} = \big[\ntkf-\ntkrf^\top\ntkr^{-1}\ntkrf\big]^{-1}$ and $\bm{V} = (\by_f-f_{\theta}(\calD_f)) + \ntkrf^\top\ntkf^{-1}(\by_r-f_{\theta}(\calD_r))$ are the re-weighting matrices, while $\by_f$ and $\by_r$ are the ground truth labels for the forget set and retain set, respectively.

Despite NTK-based unlearning showcasing state-of-the-art performance in comparison to other methods~\cite{prune_unlearn}, there are concerns regarding its numerical instability and scalability for models with many parameters \citep{golatkar2020eternal, golatkar2020forgetting}.
The inherent computational complexity has spurred efforts to enhance the efficiency of NTK-based unlearning algorithms, especially in large-scale setups.
One approach to mitigate the computational costs involves the utilization of sketching techniques to approximate tensor products associated with NTK \citep{zandieh2021scaling}. 
This method not only scales linearly with data sparsity, but also efficiently truncates the Taylor series of arc-cosine kernels.
Additionally, improvements in the spectral approximation of the kernel matrix are achieved through leveraging the score sampling, or introducing a distribution that efficiently generates random features by approximating scores of arc-cosine kernels.
Further strides in computational efficiency are made by novel algorithms employing mixed-order or high-order automatic differentiation \citep{finite_ntk}. 
It is important to note that these methods are often tailored to specific types of deep neural networks, thus limiting their widespread applicability. 
Moreover, their efficiency may still fall short for some larger deep networks~\cite{finite_ntk}. 
Consequently, our objective is to propose a parameter-efficient and practical implementation of NTK-based unlearning methods, as discussed next.

\section{Proposed Method}
The major barrier in NTK-based unlearning arises from the computation of the Jacobian matrix $\nabla_{\theta} f_\theta(\D)$, defined in Eq.~\eqref{eq:optimal-scrubbing}, with dimensions $|\calY||\calD_f|\times d$.
In the context of deep neural networks, the parameter count $d$ spans a vast range, from millions to trillions~\cite{clip,vision_transformer}.
This abundance of parameters poses a formidable challenge due to the prohibitive costs in computation and storage, and has indeed been a primary impediment in applying NTK-based unlearning algorithm on large scale models.
To mitigate the computational and storage burdens, the concept of PEFT has been proposed in Houlsby et al.~\cite{houlsby2019parameter}. 
PEFT selectively fine-tunes only a small subset of (additional) model parameters. Recent empirical findings indicate that state-of-the-art PEFT techniques achieve performance comparable to that of full fine-tuning (i.e., tuning all parameters)~\cite{zheng2022_vit_prompt}.

Drawing inspiration from PEFT, we extend the approach to NTK-based unlearning by selectively focusing on a subset of model parameters---this combined technique is referred to as ``Fast-NTK.''
As illustrated in Fig.~\ref{fig:overview}, in the case of convolutional neural networks (CNNs), our approach involves fine-tuning the batch normalization (BN) layers, which has proven to be an effective strategy for adapting the model to new data domains~\cite{chiang2023mobiletl,lin2022device_256k}.
Meanwhile, for vision transformers (ViTs), success is achieved by fine-tuning several prompts appended to the attention blocks \cite{zheng2022_vit_prompt,jia2022visual_vit_prompt,liu2023visual_vit_prompt}.
To elaborate, given a pre-trained CNN or ViT, we perform fine-tuning on the downstream dataset $\calD$ by using BN-based adjustments (for CNNs) or prompt-based modifications (for ViTs). 
Subsequently, when provided with a forget set $\calD_f$, we execute NTK-based unlearning using Eq.~\eqref{eq:optimal-scrubbing} exclusively on the fine-tuned parameters. 
This streamlined Fast-NTK approach significantly reduces the parameters subjected to fine-tuning, down to a range of $\mathbf{0.05\% \sim 4.88\%}$ of the full model parameters. Remarkably, Fast-NTK achieves a performance comparable to tuning all parameters, as demonstrated in the next section.
For an analysis on the parameter reduction, see Appendix~\ref{appendix:parameter-analysis}.

\section{Empirical results}\label{sec:exp_results}

\begin{table*}[!t]
\centering
\caption{\footnotesize BN-based \textsc{Fast-NTK} on CNNs with CIFAR-10. All metrics are averaged over 5 runs with different seeds.
}\label{tab:cnn_results}
\begin{adjustbox}{max width=1.0\textwidth}
\begin{tabular}{cccccccc}
\toprule
 & Architectures & \multicolumn{3}{c}{MobileNetV2} & \multicolumn{3}{c}{ResNet-110}  \\ \cmidrule(lr){3-5}\cmidrule(lr){6-8}
Dataset  & \#Images per class  &{100}  &{200}  & 500  &{100}  &{200}  & 500  \\ \cmidrule(lr){3-5}\cmidrule(lr){6-8}
CIFAR-10 & \#Params ratio (\%)  &{4.88}  &{4.88}  & 4.88  &{0.51}  &{0.51}  & 0.51  \\ \midrule
\multirow{7}{*}{Accuracy on $\calD_r$} & \textsc{Full} &{74.42$\pm$2.17} &{78.54$\pm$0.62} & 84.12$\pm$0.24 &{66.87$\pm$1.03} &{72.28$\pm$1.39} & 77.22$\pm$1.27 \\ \cmidrule(lr){2-8} 
 & \textsc{Max Loss} &{71.13$\pm$1.91} &{68.24$\pm$1.40} & 14.12$\pm$1.59 &{56.64$\pm$2.17} &{49.17$\pm$2.19} & 13.49$\pm$1.89 \\ \cmidrule(lr){2-8} 
 & \textsc{Random Label} &{69.58$\pm$2.21} &{69.02$\pm$1.72} & 66.94$\pm$2.84 &{58.76$\pm$1.58} &{66.58$\pm$1.71} & 72.58$\pm$2.42 \\ \cmidrule(lr){2-8} 
 & \textsc{\bf Fast-NTK} &{70.80$\pm$2.04} &{73.70$\pm$0.68} & 80.76$\pm$0.40 &{65.60$\pm$4.36} &{71.04$\pm$1.65} & 76.84$\pm$0.21 \\ \cmidrule(lr){2-8} 
 & \textsc{Retrain} &{75.56$\pm$2.36} &{79.50$\pm$0.55} & 85.27$\pm$0.26 &{69.02$\pm$1.65} &{74.13$\pm$1.54} & 78.98$\pm$0.43 \\ \midrule
\multirow{7}{*}{Accuracy on $\calD_f$} & \textsc{Full} &{68.40$\pm$5.28} &{75.00$\pm$4.17} & 84.80$\pm$2.07 &{67.20$\pm$3.06} &{73.70$\pm$1.81} & 75.20$\pm$0.98 \\ \cmidrule(lr){2-8} 
 & \textsc{Max Loss} &{0.00$\pm$0.00}  &{0.00$\pm$0.00}  & 0.00$\pm$0.00  &{0.00$\pm$0.00}  &{0.00$\pm$0.00}  & 0.00$\pm$0.00  \\ \cmidrule(lr){2-8} 
 & \textsc{Random Label} &{0.00$\pm$0.00}  &{0.00$\pm$0.00}  & 0.00$\pm$0.00  &{0.00$\pm$0.00}  &{0.00$\pm$0.00}  & 0.00$\pm$0.00  \\ \cmidrule(lr){2-8} 
 & \textsc{\bf Fast-NTK} &{0.00$\pm$0.00}  &{0.00$\pm$0.00}  & 0.00$\pm$0.00  &{0.00$\pm$0.00}  &{0.00$\pm$0.00}  & 0.00$\pm$0.00  \\ \cmidrule(lr){2-8} 
 & \textsc{Retrain} &{0.00$\pm$0.00}  &{0.00$\pm$0.00}  & 0.00$\pm$0.00  &{0.00$\pm$0.00}  &{0.00$\pm$0.00}  & 0.00$\pm$0.00  \\ \midrule
\multirow{7}{*}{{\makecell[c]{Accuracy\\on Hold-Out set}}}  & \textsc{Full} &{65.00$\pm$1.11} &{71.63$\pm$1.25} & 77.91$\pm$0.21 &{54.12$\pm$0.72} &{62.29$\pm$1.12} & 71.02$\pm$0.71 \\ \cmidrule(lr){2-8} 
 & \textsc{Max Loss} &{58.14$\pm$0.95} &{58.28$\pm$1.22} & 12.51$\pm$1.27 &{43.18$\pm$0.34} &{41.61$\pm$2.06} & 12.06$\pm$2.81 \\ \cmidrule(lr){2-8} 
 & \textsc{Random Label} &{57.04$\pm$1.15} &{63.36$\pm$0.48} & 69.27$\pm$0.15 &{43.38$\pm$0.91} &{52.53$\pm$0.64} & 61.18$\pm$0.60 \\ \cmidrule(lr){2-8} 
 & \textsc{\bf Fast-NTK} &{58.54$\pm$0.88} &{63.96$\pm$1.67} & 69.96$\pm$3.64 &{50.80$\pm$5.57} &{59.88$\pm$1.59} & 60.58$\pm$0.63 \\ \cmidrule(lr){2-8} 
 & \textsc{Retrain} &{60.50$\pm$1.15} &{66.41$\pm$0.46} & 71.80$\pm$0.19 &{50.36$\pm$1.17} &{57.57$\pm$0.71} & 65.58$\pm$1.51 \\ \midrule
\multirow{6}{*}{{\makecell[c]{\#Relearning\\Epochs}}} & \textsc{Max Loss} &{28.80$\pm$0.40} &{22.20$\pm$0.40} & 77.20$\pm$6.01 &{24.00$\pm$0.89} &{25.20$\pm$2.48} & 22.00$\pm$0.93 \\ \cmidrule(lr){2-8} 
 & \textsc{Random Label} &{19.80$\pm$0.40} &{10.00$\pm$0.00} & 4.00$\pm$0.00  &{10.80$\pm$0.40} &{6.00$\pm$0.00}  & 3.00$\pm$0.49  \\ \cmidrule(lr){2-8} 
 & \textsc{\bf Fast-NTK} &{21.00$\pm$0.63} &{10.80$\pm$0.40} & 4.00$\pm$0.00  &{12.40$\pm$0.80} &{6.00$\pm$0.00}  & 2.80$\pm$0.40  \\ \cmidrule(lr){2-8} 
 & \textsc{Retrain} &{21.20$\pm$0.40} &{11.00$\pm$0.00} & 4.80$\pm$0.40  &{12.60$\pm$0.49} &{6.20$\pm$0.40}  & 3.00$\pm$0.00  \\ \bottomrule
\end{tabular}\end{adjustbox}
\end{table*}

\subsection{Setup}\label{sec:main_setup}
Our method starts with the CNNs and ViTs pre-trained on the CIFAR-100 and ImageNet-1K datasets respectively. 
We fine-tune these models on the CIFAR-10 \cite{krizhevsky2009learning} and ImageNet-R \cite{hendrycks2020many} datasets and then assess the performance of  \textsc{Fast-NTK}. 
In the case of CIFAR-10, we designate one class as $\calD_f$, while considering the remaining classes as $\calD_r$. 
Similarly, for the ImageNet-R dataset, we randomly choose one class as $\calD_f$ and select either 19 or 49 classes from the 200 classes as $\calD_r$ (i.e., resulting in 20 or 50 total classes in $\calD$) to demonstrate the scalability of our approach.

We consider the following three metrics. 
First, we measure accuracy on both $\calD_r$ and $\calD_f$---an unlearning algorithm should maintain high accuracy on $\calD_r$ while minimizing accuracy on $\calD_f$. 
Second, we calculate accuracy on a hold-out set to ensure consistent performance on unseen data. 
Note that the hold-out set may contain samples from classes present in both $\calD_f$ and $\calD_r$. 
The accuracy on the hold-out set should remain unaffected by the unlearning algorithm.
Third, we incorporate relearning time \cite{golatkar2020eternal}, representing the number of epochs required to achieve a training loss below 0.05 on the forget set.
Here, the value 0.05 is manually chosen, and could be chosen to other values. 
Relearning time serves as a measure of the difficulty in recovering knowledge from the forget set. 
If the model fails to achieve a loss below 0.05 within 100 epochs, we denote it as `$>$100'.

We compare \textsc{Fast-NTK} against the following baselines:
\begin{itemize}
  \item \textsc{Full}: The original model fine-tuned on $\calD = \calD_f \cup \calD_r$ without unlearning, serving as the reference model.
  \item \textsc{Max Loss} \cite{halimi2022federated}: This baseline maximizes the training loss with respect to the ground truth labels of the samples in the forget set $\calD_f$.
  \item \textsc{Random Label} \cite{graves2021amnesiac,mul_generative3}: This baseline minimizes the training loss by assigning uniformly random labels to the samples in the forget set $\calD_f$.
  \item \textsc{Retrain}: The model trained only on the retain set $\calD_r$.
\end{itemize}

Among these baselines, \textsc{Retrain} is commonly referred to as the \textbf{golden baseline}. 
This designation stems from its lack of prior knowledge about the samples in the forget set $\calD_f$, making it an ideal reference point for comparing any unlearning algorithms. 
By evaluating \textsc{Fast-NTK} against \textsc{Retrain}, we aim to ensure that the unlearned model closely approximates the ideal scenario. 
This comparison helps ascertain that the unlearning process effectively eliminates unwanted data without causing significant performance degradation on $\calD_r$. 
Essentially, an ideal unlearned model should exhibit indistinguishability in terms of the specified evaluation metrics to the golden baseline \textsc{Retrain} (see~\cite[Section~3.2]{unlearn_survey}).

\begin{table*}[t!]
\centering
\caption{\footnotesize 
Prompt-based \textsc{Fast-NTK} on ViTs with ImageNet-R. All metrics are averaged over 5 runs with different seeds. 
}\label{tab:vit_results}
\begin{adjustbox}{max width=\textwidth}
\begin{tabular}{cccccccc}
\toprule
{} & {Architectures} & \multicolumn{2}{c}{{ViT-Tiny}}  & \multicolumn{2}{c}{{ViT-Small}}  & \multicolumn{2}{c}{{ViT-Base}}  \\ \cmidrule(lr){3-4}\cmidrule(lr){5-6}\cmidrule(lr){7-8}
{Dataset} & {\#Classes/\#IPC}  & {{20/50}}  & {50/20} & {{20/50}}  & {50/20}  & {{20/50}} & {50/20}  \\ \cmidrule(lr){3-4}\cmidrule(lr){5-6}\cmidrule(lr){7-8}
{ImageNet-R} & {\#Params ratio (\%)}  & {0.24}  & {0.35} & {0.12}  & {0.18}  & {0.06} & {0.09}  \\ \midrule
{} & {\textsc{Full}} & {{66.40$\pm$0.91}} & {65.48$\pm$0.85}  & {{87.60$\pm$0.89}} & {85.56$\pm$1.58} & {{36.82$\pm$2.55}}  & {15.36$\pm$1.17} \\ \cmidrule(lr){2-8} 
{} & {\textsc{Max Loss}} & {{57.71$\pm$1.33}} & {51.80$\pm$0.70}  & {{77.35$\pm$1.23}} & {71.17$\pm$0.66} & {{24.78$\pm$2.74}}  & {8.52$\pm$0.77}  \\ \cmidrule(lr){2-8} 
{} & {\textsc{Random Label}} & {{58.29$\pm$1.70}} & {51.50$\pm$0.97}  & {{78.51$\pm$1.52}} & {71.43$\pm$0.20} & {{23.56$\pm$2.99}}  & {7.60$\pm$0.77}  \\ \cmidrule(lr){2-8} 
{} & {\textsc{\bf Fast-NTK}} & {{66.53$\pm$0.63}} & {65.24$\pm$0.48}  & {{87.03$\pm$1.49}} & {85.31$\pm$2.14} & {{40.84$\pm$1.84}}  & {17.40$\pm$1.89} \\ \cmidrule(lr){2-8} 
\multirow{-7}{*}{{Accuracy on $\calD_r$}} & {\textsc{Retrain}}  & {{68.21$\pm$1.50}} & {65.92$\pm$0.93}  & {{87.77$\pm$0.42}} & {86.58$\pm$0.15} & {{37.45$\pm$2.03}}  & {16.02$\pm$1.02} \\ \midrule
{} & {\textsc{Full}} & {{77.20$\pm$5.60}} & {56.67$\pm$20.95} & {{91.60$\pm$3.44}} & {87.50$\pm$2.50} & {{56.80$\pm$11.91}} & {20.00$\pm$5.00} \\ \cmidrule(lr){2-8} 
{} & {\textsc{Max Loss}} & {{0.00$\pm$0.00}}  & {0.00$\pm$0.00} & {{0.00$\pm$0.00}}  & {0.00$\pm$0.00}  & {{0.00$\pm$0.00}} & {0.00$\pm$0.00}  \\ \cmidrule(lr){2-8} 
{} & {\textsc{Random Label}} & {{0.00$\pm$0.00}} & {0.00$\pm$0.00} & {{0.00$\pm$0.00}} & {0.00$\pm$0.00} & {{0.00$\pm$0.00}} & {0.00$\pm$0.00}  \\ \cmidrule(lr){2-8} 
{} & {\textsc{\bf Fast-NTK}} & {{0.00$\pm$0.00}}  & {0.00$\pm$0.00} & {{0.00$\pm$0.00}}  & {3.00$\pm$2.50}  & {{0.00$\pm$0.00}} & {0.00$\pm$0.00}  \\ \cmidrule(lr){2-8} 
\multirow{-7}{*}{{Accuracy on $\calD_f$}} & {\textsc{Retrain}}  & {{0.00$\pm$0.00}}  & {0.00$\pm$0.00} & {{0.00$\pm$0.00}}  & {0.00$\pm$0.00}  & {{0.00$\pm$0.00}} & {0.00$\pm$0.00}  \\ \midrule
{} & {\textsc{Full}} & {{47.73$\pm$0.45}} & {31.43$\pm$1.70}  & {{68.06$\pm$1.12}} & {52.75$\pm$0.75} & {{32.53$\pm$2.40}}  & {12.05$\pm$0.05} \\ \cmidrule(lr){2-8} 
{} & {\textsc{Max Loss}} & {{41.56$\pm$1.05}} & {26.13$\pm$0.87}  & {{59.41$\pm$1.03}} & {45.15$\pm$0.05} & {{19.67$\pm$3.20}}  & {6.55$\pm$0.55}  \\ \cmidrule(lr){2-8} 
{} & {\textsc{Random Label}} & {{45.02$\pm$1.92}} & {26.03$\pm$1.31}  & {{64.03$\pm$1.23}} & {45.40$\pm$0.10} & {{19.96$\pm$3.60}}  & {6.55$\pm$0.25}  \\ \cmidrule(lr){2-8} 
{} & {\textsc{\bf Fast-NTK}} & {{45.44$\pm$0.93}} & {31.17$\pm$1.28}  & {{64.03$\pm$1.03}} & {52.40$\pm$0.50} & {{23.26$\pm$2.40}}  & {9.15$\pm$0.35}  \\ \cmidrule(lr){2-8} 
\multirow{-7}{*}{{\makecell[c]{Accuracy\\on Hold-Out set}}} & {\textsc{Retrain}}  & {{46.54$\pm$1.26}} & {30.67$\pm$2.25}  & {{64.69$\pm$0.90}} & {51.60$\pm$0.90} & {{30.29$\pm$2.24}}  & {11.60$\pm$0.10} \\ \midrule
{} & {\textsc{Max Loss}} & {{17.00$\pm$0.00}} & {13.67$\pm$0.47}  & {{18.00$\pm$0.00}} & {15.00$\pm$0.00} & {{$>$100}}  & {$>$100} \\ \cmidrule(lr){2-8} 
{} & {\textsc{Random Label}} & {{4.20$\pm$0.40}}  & {3.67$\pm$0.47} & {{6.40$\pm$0.49}}  & {6.00$\pm$0.00}  & {{$>$100}}  & {$>$100} \\ \cmidrule(lr){2-8} 
{} & {\textsc{\bf Fast-NTK}} & {{4.40$\pm$0.49}}  & {4.00$\pm$0.00} & {{5.80$\pm$0.40}}  & {6.00$\pm$0.00}  & {{$>$100}}  & {$>$100} \\ \cmidrule(lr){2-8} 
\multirow{-6}{*}{{\makecell[c]{\#Relearning\\Epochs}}}  & {\textsc{Retrain}}  & {{5.00$\pm$0.00}}  & {4.67$\pm$0.47} & {{6.40$\pm$0.49}}  & {6.50$\pm$0.50}  & {{$>$100}}  & {$>$100} \\ \bottomrule
\end{tabular}
\end{adjustbox}
\end{table*}

\begin{table*}[!thb]
\centering
\caption{\footnotesize 
Linear probing on the ImageNet-R dataset. All metrics are averaged over 5 runs with different seeds. 
}\label{tab:linearprobe}
\begin{tabular}{cccccccc}
\toprule
  & {Network} & \multicolumn{3}{c}{ViT-Small}  & \multicolumn{3}{c}{ViT-Base}  \\ \cmidrule(lr){3-5}\cmidrule(lr){6-8}  
  & {\#Classes/\#IPC} & {20/20} & {20/50} & {50/20} & {20/20} & {20/50}  & {50/20} \\ 
  \midrule
\multirow{4}{*}{Acc on $\calD_r$} & \textsc{Pre-Trained}  & {60.39$\pm$1.27} & {58.24$\pm$1.49}  & {53.70$\pm$2.36}  & {99.93$\pm$0.11} & {99.32$\pm$0.24} & {99.87$\pm$0.08}\\ \cmidrule(lr){2-8} 
  &\textsc{Random-Init}    & {35.66$\pm$1.69} & {26.40$\pm$1.06}  & 22.32$\pm$0.60& {32.31$\pm$0.74} & {19.30$\pm$0.66} & 17.33$\pm$0.69  \\ \cmidrule(lr){2-8} 
  &\textsc{\bf Fast-NTK}  & 60.25$\pm$3.76 & 53.71$\pm$2.45 & 47.24$\pm$0.00  & 86.58$\pm$2.13 & 87.66$\pm$1.01 & 87.24$\pm$0.00  \\
  \midrule
\multirow{4}{*}{Acc on $\calD_f$} & \textsc{Pre-Trained}   & {72.50$\pm$9.01} & {79.50$\pm$6.22}  & {66.25$\pm$5.45} & {100.00$\pm$0.00}  & {99.00$\pm$1.00} & {98.75$\pm$2.17} \\ \cmidrule(lr){2-8} 
  &\textsc{Random-Init}    & {54.53$\pm$2.46} & {33.33$\pm$17.00} & 43.33$\pm$11.12 & {49.40$\pm$4.42} & {15.00$\pm$8.16} & 17.33$\pm$7.72 \\ \cmidrule(lr){2-8} 
  &\textsc{\bf Fast-NTK}   & {0.00$\pm$0.00}  & {0.00$\pm$0.00} & 0.00$\pm$0.00& {2.50$\pm$2.50}  & {0.00$\pm$0.00}  & 0.00$\pm$0.00 \\
  \bottomrule
\end{tabular}
\end{table*}

\subsection{Evaluation of Fast-NTK}
We perform BN-based fine-tuning on MobileNet-v2 and ResNet-110 using a subset of the CIFAR-10 dataset, followed by unlearning algorithms that involves forgetting the class labeled ``0.'' 
To showcase the scalability of our approach, we vary the number of images per class (\#IPC). 
The results in Table~\ref{tab:cnn_results} reveal that our method requires less than {\bf 4.88\%} of the parameters involved in tuning the entire model, making the unlearning process practical and achievable for these large models. 
Notably, \textsc{Fast-NTK} exhibits negligible or no accuracy degradation on the retain set compared to the golden baseline \textsc{Retrain}. 
In contrast, the accuracy on the forget set is indistinguishable from \textsc{Retrain} (drops to "0") across various setups, with a similar number of relearning epochs needed as \textsc{Retrain}. 
Compared to the other baselines, \textsc{Max Loss} and \textsc{Random Label}, \textsc{Fast-NTK} effectively preserves accuracy on the retrain set $\calD_r$ and the general test set, highlighting the robustness and efficiency of our proposed technique for CNNs.

Additionally, we extend the same setting to ViTs on the ImageNet-R dataset.
As demonstrated in Table~\ref{tab:vit_results}, our approach requires less than {\bf 0.4\%} of the parameters compared to tuning the entire model, making practical unlearning feasible for these large models. 
Comparisons with \textsc{Retrain}, \textsc{Max Loss}, and \textsc{Random Label} show that \textsc{Fast-NTK} effectively preserves accuracy on the retain set $\calD_r$ and the general test set, achieving close accuracy to \textsc{Retrain} on the retain set. 
These results confirm the effectiveness and practicality of our unlearning approach for ViTs. 
Importantly, our method scales up to ViTs, representing a significant advancement compared to previous approaches like \cite{golatkar2020eternal}, which are confined only to toy networks and small datasets (e.g., less than 200 samples). 
For additional results on CIFAR-10, see Appendix~\ref{appendix:additional-results}.

\section{Discussion}
\paragraph{Risk of using pre-trained models.}
It is crucial to emphasize that \textsc{Fast-NTK} starts with a pre-trained model rather than one initialized randomly. 
Despite the increasing popularity of leveraging pre-trained foundation models~\cite{bommasani2021opportunities}, these pre-trained models may possess some knowledge of classes from $\calD_f$. 
This prior knowledge introduces an inherent risk during the unlearning process, as erasing all information and concepts associated with the classes in $\calD_f$ solely through the use of forget samples becomes a challenging task.
To reassess this risk, for the pre-trained models used in our evaluation (\textsc{Pre-Trained}), we conduct fine-tuning of the classification head (i.e., linear probing) on $\calD_r \cup \calD_f$, while keeping the parameters in the remaining layers frozen. 
We also conduct the linear probing on the randomly initialized model (\textsc{Random-Init}) and the unlearned model obtained by \textsc{Fast-NTK} (cf. Section~\ref{sec:exp_results}). 
As illustrated in Table~\ref{tab:linearprobe}, the accuracy of \textsc{Pre-Trained} on $\calD_r$ and $\calD_f$ is much higher than \textsc{Random-Init} (very close to 100\%), indicating that the pre-trained model already possesses some level of knowledge about $\calD_r$ and $\calD_f$.
As expected, \textsc{Fast-NTK} effectively removes the knowledge on $\calD_f$ as the accuracy on $\calD_f$ is zero. 
This finding underscores the need for further investigation into the interplay between unlearning and PEFT on pre-trained models.

\paragraph{Future work.} 
Our current implementation to obtain the NTK matrix relies on exact computations. 
To further improve the efficiency of \textsc{Fast-NTK}, one future direction is to explore approximate computation of the NTK matrix by, e.g., low-rank approximation or factorization.

\clearpage 
{\footnotesize
\section*{Disclaimer}
This paper was prepared for informational purposes by the Global Technology Applied Research center of JPMorgan Chase \& Co. This paper is not a product of the Research Department of JPMorgan Chase \& Co. or its affiliates. Neither JPMorgan Chase \& Co. nor any of its affiliates makes any explicit or implied representation or warranty and none of them accept any liability in connection with this paper, including, without limitation, with respect to the completeness, accuracy, or reliability of the information contained herein and the potential legal, compliance, tax, or accounting effects thereof. This document is not intended as investment research or investment advice, or as a recommendation, offer, or solicitation for the purchase or sale of any security, financial instrument, financial product or service, or to be used in any way for evaluating the merits of participating in any transaction. Guihong Li's and Radu Marculescu's contributions were made as part of Guihong Li's internship at the Global Technology Applied Research center of JPMorgan Chase \& Co.
}

\bibliographystyle{IEEEtran}
\bibliography{reference.bib}

\begin{thebibliography}{10}
\providecommand{\url}[1]{#1}
\csname url@samestyle\endcsname
\providecommand{\newblock}{\relax}
\providecommand{\bibinfo}[2]{#2}
\providecommand{\BIBentrySTDinterwordspacing}{\spaceskip=0pt\relax}
\providecommand{\BIBentryALTinterwordstretchfactor}{4}
\providecommand{\BIBentryALTinterwordspacing}{\spaceskip=\fontdimen2\font plus
\BIBentryALTinterwordstretchfactor\fontdimen3\font minus
  \fontdimen4\font\relax}
\providecommand{\BIBforeignlanguage}[2]{{%
\expandafter\ifx\csname l@#1\endcsname\relax
\typeout{** WARNING: IEEEtran.bst: No hyphenation pattern has been}%
\typeout{** loaded for the language `#1'. Using the pattern for}%
\typeout{** the default language instead.}%
\else
\language=\csname l@#1\endcsname
\fi
#2}}
\providecommand{\BIBdecl}{\relax}
\BIBdecl

\bibitem{eu_gdpr}
G.~D.~P. Regulation, ``General data protection regulation (gdpr),''
  \emph{Intersoft Consulting, Accessed in October}, vol.~24, no.~1, 2018.

\bibitem{iclr_gan_memory}
Y.~Wu, Y.~Burda, R.~Salakhutdinov, and R.~B. Grosse, ``On the quantitative
  analysis of decoder-based generative models,'' in \emph{Proceedings of ICLR},
  2017.

\bibitem{unlearn_survey}
T.~T. Nguyen, T.~T. Huynh, P.~L. Nguyen, A.~W. Liew, H.~Yin, and Q.~V.~H.
  Nguyen, ``A survey of machine unlearning,'' \emph{CoRR}, vol. abs/2209.02299,
  2022.

\bibitem{sisa_unlearn}
L.~Bourtoule, V.~Chandrasekaran, C.~A. Choquette{-}Choo, H.~Jia, A.~Travers,
  B.~Zhang, D.~Lie, and N.~Papernot, ``Machine unlearning,'' in \emph{42nd
  {IEEE} Symposium on Security and Privacy}.\hskip 1em plus 0.5em minus
  0.4em\relax {IEEE}, 2021, pp. 141--159.

\bibitem{gupta2021adaptive}
V.~Gupta, C.~Jung, S.~Neel, A.~Roth, S.~Sharifi-Malvajerdi, and C.~Waites,
  ``Adaptive machine unlearning,'' in \emph{Advances in NeurIPS}, 2021.

\bibitem{neel2021descent}
S.~Neel, A.~Roth, and S.~Sharifi{-}Malvajerdi, ``Descent-to-delete:
  Gradient-based methods for machine unlearning,'' in \emph{Proceedings of
  ALT}.\hskip 1em plus 0.5em minus 0.4em\relax {PMLR}, 2021.

\bibitem{tarun2023deep}
A.~K. Tarun, V.~S. Chundawat, M.~Mandal, and M.~S. Kankanhalli, ``Deep
  regression unlearning,'' in \emph{Proceedings of ICML}.\hskip 1em plus 0.5em
  minus 0.4em\relax {PMLR}, 2023.

\bibitem{grad_up}
R.~Chourasia and N.~Shah, ``Forget unlearning: Towards true data-deletion in
  machine learning,'' in \emph{Proceedings of ICML}.\hskip 1em plus 0.5em minus
  0.4em\relax {PMLR}, 2023.

\bibitem{chen2019novel}
Y.~Chen, J.~Xiong, W.~Xu, and J.~Zuo, ``A novel online incremental and
  decremental learning algorithm based on variable support vector machine,''
  \emph{Cluster Computing}, vol.~22, pp. 7435--7445, 2019.

\bibitem{golatkar2020forgetting}
A.~Golatkar, A.~Achille, and S.~Soatto, ``Forgetting outside the box: Scrubbing
  deep networks of information accessible from input-output observations,'' in
  \emph{Proceedings of ECCV}.\hskip 1em plus 0.5em minus 0.4em\relax Springer,
  2020.

\bibitem{golatkar2020eternal}
------, ``Eternal sunshine of the spotless net: Selective forgetting in deep
  networks,'' in \emph{Proceedings of CVPR}.\hskip 1em plus 0.5em minus
  0.4em\relax {IEEE}, 2020.

\bibitem{lin2022device_256k}
J.~Lin, L.~Zhu, W.-M. Chen, W.-C. Wang, C.~Gan, and S.~Han, ``On-device
  training under 256kb memory,'' in \emph{Advances in NeurIPS}, 2022.

\bibitem{zheng2022_vit_prompt}
Z.~Zheng, X.~Yue, K.~Wang, and Y.~You, ``Prompt vision transformer for domain
  generalization,'' \emph{arXiv preprint arXiv:2208.08914}, 2022.

\bibitem{jia2022visual_vit_prompt}
M.~Jia, L.~Tang, B.-C. Chen, C.~Cardie, S.~Belongie, B.~Hariharan, and S.-N.
  Lim, ``Visual prompt tuning,'' in \emph{Proceedings of ECCV}.\hskip 1em plus
  0.5em minus 0.4em\relax Springer, 2022.

\bibitem{chiang2023mobiletl}
H.-Y. Chiang, N.~Frumkin, F.~Liang, and D.~Marculescu, ``Mobile{TL}: on-device
  transfer learning with inverted residual blocks,'' in \emph{Proceedings of
  the AAAI}, 2023.

\bibitem{mul_survey2023}
H.~Xu, T.~Zhu, L.~Zhang, W.~Zhou, and P.~S. Yu, ``Machine unlearning: A
  survey,'' \emph{ACM Comput. Surv.}, vol.~56, no.~1, aug 2023.

\bibitem{lee2019wide}
J.~Lee \emph{et~al.}, ``Wide neural networks of any depth evolve as linear
  models under gradient descent,'' in \emph{Advances in NeurIPS}, 2019.

\bibitem{jacot2018neural}
A.~Jacot, F.~Gabriel, and C.~Hongler, ``Neural tangent kernel: Convergence and
  generalization in neural networks,'' in \emph{Advances in NeurIPS}, 2018.

\bibitem{prune_unlearn}
J.~Jia, J.~Liu, P.~Ram, Y.~Yao, G.~Liu, Y.~Liu, P.~Sharma, and S.~Liu, ``Model
  sparsification can simplify machine unlearning,'' \emph{CoRR}, vol.
  abs/2304.04934, 2023.

\bibitem{zandieh2021scaling}
A.~Zandieh, I.~Han, H.~Avron, N.~Shoham, C.~Kim, and J.~Shin, ``Scaling neural
  tangent kernels via sketching and random features,'' in \emph{Advances in
  NeurIPS}, 2021.

\bibitem{finite_ntk}
R.~Novak, J.~Sohl{-}Dickstein, and S.~S. Schoenholz, ``Fast finite width neural
  tangent kernel,'' in \emph{Proceedings of ICML}.\hskip 1em plus 0.5em minus
  0.4em\relax {PMLR}, 2022.

\bibitem{clip}
A.~Radford, J.~W. Kim, C.~Hallacy, A.~Ramesh, G.~Goh, S.~Agarwal, G.~Sastry,
  A.~Askell, P.~Mishkin, J.~Clark, G.~Krueger, and I.~Sutskever, ``Learning
  transferable visual models from natural language supervision,'' in
  \emph{Proceedings of ICML}.\hskip 1em plus 0.5em minus 0.4em\relax {PMLR},
  2021.

\bibitem{vision_transformer}
A.~Dosovitskiy \emph{et~al.}, ``An image is worth 16x16 words: Transformers for
  image recognition at scale,'' in \emph{Proceedings of ICLR}, 2021.

\bibitem{houlsby2019parameter}
N.~Houlsby, A.~Giurgiu, S.~Jastrzebski, B.~Morrone, Q.~De~Laroussilhe,
  A.~Gesmundo, M.~Attariyan, and S.~Gelly, ``Parameter-efficient transfer
  learning for nlp,'' in \emph{Proceedings of ICML}.\hskip 1em plus 0.5em minus
  0.4em\relax PMLR, 2019.

\bibitem{liu2023visual_vit_prompt}
H.~Liu, C.~Li, Q.~Wu, and Y.~J. Lee, ``Visual instruction tuning,'' \emph{arXiv
  preprint arXiv:2304.08485}, 2023.

\bibitem{krizhevsky2009learning}
A.~Krizhevsky, G.~Hinton \emph{et~al.}, ``Learning multiple layers of features
  from tiny images,'' 2009.

\bibitem{hendrycks2020many}
D.~Hendrycks \emph{et~al.}, ``The many faces of robustness: A critical analysis
  of out-of-distribution generalization,'' \emph{arXiv preprint
  arXiv:2006.16241}, 2020.

\bibitem{halimi2022federated}
A.~Halimi, S.~Kadhe, A.~Rawat, and N.~Baracaldo, ``Federated unlearning: How to
  efficiently erase a client in fl?'' \emph{CoRR}, vol. abs/2207.05521, 2022.

\bibitem{graves2021amnesiac}
L.~Graves, V.~Nagisetty, and V.~Ganesh, ``Amnesiac machine learning,'' in
  \emph{Proceedings of the AAAI}, 2021, pp. 11\,516--11\,524.

\bibitem{mul_generative3}
Z.~Kong and K.~Chaudhuri, ``Data redaction from conditional generative
  models,'' \emph{CoRR}, vol. abs/2305.11351, 2023.

\bibitem{bommasani2021opportunities}
R.~Bommasani \emph{et~al.}, ``On the opportunities and risks of foundation
  models,'' \emph{arXiv preprint arXiv:2108.07258}, 2021.

\end{thebibliography}

\clearpage 
\onecolumn
\appendix 
\setcounter{table}{0}
\def\theequation{A.\arabic{equation}}
\def\thetable{A.\Roman{table}}

\subsection{Parameter Reduction of Fast-NTK}\label{appendix:parameter-analysis}
\paragraph{Fine-tune/unlearn CNNs with BN layers.}
As shown in Fig.~\ref{fig:overview}, a convolutional layer is typically followed by a batch normalization layer in a CNN.
For a typical convolutional layer with \( C_o \) output channels, \( C_i \) input channels, kernel size \( K\times K \),  and \( g \) separable groups, the total number of parameters (weights) in this layer is
\(\text{Parameters}_{\text{conv}} = \frac{C_o \times C_i \times K^2}{g}\). 
In contrast, for a batch normalization (BN) layer, the only learnable parameters are the scaling (\( \boldsymbol{\gamma} \)) and shifting (\( \boldsymbol{\beta} \)) 
terms for each channel. Hence, the total number of learnable parameters in a BN layer is then $ \text{Parameters}_{\text{BN}} = 2 \times C_o$. 
Usually, $C_i \times K^2 \gg 2$ and $g = 1$; therefore 
\begin{equation}
\frac{\text{Parameters}_{\text{conv}}}{\text{Parameters}_{\text{BN}}}=\frac{ C_i \times K^2}{2g} \gg 1.
\end{equation}

\paragraph{Fine-tune/unlearn ViTs with Prompts.}
In a ViT, the embedding layer transforms the input image into a sequence-like feature representation  with the embedding dimension of $E$. 
Then the representation is processed by several transformer block consisting of a multi-head self-attention (MSA) block and two multi-layer perceptron (MLP) layers to obtain the outputs. 
Within each block, each MLP layer has $ E\times rE$, where $r$ is usually 4; so two MLP layers have $8E^2$ parameters. Besides, each MSA has three weight matrices of size \( \frac{E}{m} \times E \).
Hence, MSA has \(3E^2\) parameters, and in total, \(\text{Parameters}_{\text{Block}} = 11E^2\)
As shown in Fig.~\ref{fig:overview}, the prompt-based fine-tuning inserts the prompt parameters $\vp_K$ and $\vp_V$ to the Key and Value $\vh_K$ and $\vh_V$ of an MSA.  
As a contrast to tuning the entire MSA, the prompt-based method fine-tunes only ${L_{p} \times E}$ parameters. 
Typically, the embedding dimensions $E$ is much higher than the prompt length $L_p$ (in our experimental setup, $L_p=10$); therefore:
\begin{equation}
\frac{\text{Parameters}_{\text{MSA}}}{\text{Parameters}_{\text{prompt}}} = \frac{11E}{L_p}\gg 1.
\end{equation}

\subsection{Additional Results on CIFAR-10.}\label{appendix:additional-results}
We provide the results of Fast-NTK with ViTs on the CIFAR-10 dataset.
Again, our approach requires less than 0.2\% of the parameters and outperforms other baselines.

\begin{table*}[htb]
\centering
\caption{
Prompt-based \textsc{Fast-NTK} on ViTs with CIFAR-10. All metrics are averaged over 5 runs with different seeds.
}\label{tab:more_vit_results}
\begin{adjustbox}{max width=1.\textwidth}
\begin{tabular}{cccccccc}
\toprule
  & Architectures & \multicolumn{3}{c}{ViT-Small} & \multicolumn{3}{c}{ViT-Base}  \\ \cmidrule(lr){3-5}\cmidrule(lr){6-8} 
Dataset  & \#Images per class  & {100}  & {200}  & 500  & {100} & {200} & 500  \\ \cmidrule(lr){3-5}\cmidrule(lr){6-8} 
CIFAR-10  & \#Params ratio (\%)  & {0.11}  & {0.11}  & 0.11  & {0.05} & {0.05} & 0.05  \\ \midrule
\multirow{7}{*}{Accuracy on $\calD_r$} & \textsc{Full} & {95.78$\pm$0.52} & {94.93$\pm$1.06} & 94.78$\pm$0.48 & {84.18$\pm$1.09}  & {85.67$\pm$0.62}  & 87.07$\pm$0.24 \\ \cmidrule(lr){2-8} 
  & \textsc{Max Loss} & {87.18$\pm$1.19} & {86.53$\pm$0.79} & 83.47$\pm$0.40 & {78.04$\pm$0.60}  & {84.26$\pm$0.83}  & 87.39$\pm$0.08 \\ \cmidrule(lr){2-8} 
  & \textsc{Random Label} & {93.87$\pm$0.86} & {93.72$\pm$0.55} & 93.32$\pm$0.35 & {76.56$\pm$0.83}  & {83.83$\pm$0.82}  & 87.28$\pm$0.17 \\ \cmidrule(lr){2-8} 
  & \textsc{\bf Fast-NTK} & {93.91$\pm$0.77} & {94.84$\pm$1.25} & 94.59$\pm$0.03 & {87.60$\pm$1.16}  & {89.13$\pm$0.51}  & 89.30$\pm$0.12 \\ \cmidrule(lr){2-8} 
  & \textsc{Retrain} & {96.02$\pm$0.43} & {95.71$\pm$0.53} & 94.29$\pm$0.20 & {84.36$\pm$1.16}  & {86.47$\pm$0.58}  & 88.19$\pm$0.32 \\ \midrule
\multirow{7}{*}{Accuracy on $\calD_f$} & \textsc{Full} & {97.00$\pm$1.55} & {96.20$\pm$1.36} & 95.73$\pm$1.67 & {84.80$\pm$6.31}  & {90.40$\pm$1.11}  & 92.00$\pm$0.00 \\ \cmidrule(lr){2-8} 
  & \textsc{Max Loss} & {0.00$\pm$0.00}  & {0.00$\pm$0.00}  & 0.00$\pm$0.00  & {0.00$\pm$0.00} & {0.00$\pm$0.00} & 0.00$\pm$0.00  \\ \cmidrule(lr){2-8} 
  & \textsc{Random Label} & {0.00$\pm$0.00}  & {0.00$\pm$0.00}  & 0.00$\pm$0.00  & {0.00$\pm$0.00} & {0.00$\pm$0.00} & 0.00$\pm$0.00  \\ \cmidrule(lr){2-8} 
  & \textsc{\bf Fast-NTK} & {0.20$\pm$0.40}  & {0.20$\pm$0.24}  & 0.00$\pm$0.00  & {0.00$\pm$0.00} & {0.00$\pm$0.00} & 0.20$\pm$0.20  \\ \cmidrule(lr){2-8} 
  & \textsc{Retrain} & {0.00$\pm$0.00}  & {0.00$\pm$0.00}  & 0.00$\pm$0.00  & {0.00$\pm$0.00} & {0.00$\pm$0.00} & 0.00$\pm$0.00  \\ \midrule
\multirow{7}{*}{\makecell[c]{Accuracy\\on Hold-Out set}} & \textsc{Full} & {86.62$\pm$1.42} & {87.51$\pm$0.93} & 89.29$\pm$0.18 & {82.06$\pm$0.77}  & {84.95$\pm$0.95}  & 86.78$\pm$0.28 \\ \cmidrule(lr){2-8} 
  & \textsc{Max Loss} & {71.90$\pm$0.45} & {73.38$\pm$0.72} & 73.09$\pm$0.24 & {68.14$\pm$0.71}  & {74.60$\pm$0.86}  & 77.73$\pm$0.21 \\ \cmidrule(lr){2-8} 
  & \textsc{Random Label} & {78.12$\pm$0.91} & {79.01$\pm$0.59} & 80.30$\pm$0.07 & {66.80$\pm$1.23}  & {74.34$\pm$0.83}  & 77.73$\pm$0.35 \\ \cmidrule(lr){2-8} 
  & \textsc{\bf Fast-NTK} & {77.78$\pm$1.14} & {78.87$\pm$0.97} & 80.63$\pm$0.23 & {70.62$\pm$2.10}  & {75.37$\pm$0.45}  & 78.47$\pm$0.15 \\ \cmidrule(lr){2-8} 
  & \textsc{Retrain} & {78.94$\pm$1.11} & {79.61$\pm$0.44} & 80.64$\pm$0.10 & {73.76$\pm$0.43}  & {76.56$\pm$0.87}  & 78.59$\pm$0.27 \\ \midrule
\multirow{6}{*}{\makecell[c]{\#Relearning\\Epochs}}  & \textsc{Max Loss} & {9.20$\pm$0.40}  & {8.00$\pm$0.00}  & 6.00$\pm$0.00  & {\textgreater{}100} & {\textgreater{}100} & 47.50$\pm$0.50 \\ \cmidrule(lr){2-8} 
  & \textsc{Random Label} & {2.20$\pm$0.40}  & {1.20$\pm$0.40}  & 1.00$\pm$0.00  & {\textgreater{}100} & {\textgreater{}100} & 45.00$\pm$1.00 \\ \cmidrule(lr){2-8} 
  & \textsc{\bf Fast-NTK} & {2.60$\pm$0.49}  & {1.00$\pm$0.00}  & 1.00$\pm$0.00  & {\textgreater{}100} & {\textgreater{}100} & 53.50$\pm$1.50 \\ \cmidrule(lr){2-8} 
  & \textsc{Retrain} & {2.60$\pm$0.49}  & {1.40$\pm$0.49}  & 1.00$\pm$0.00  & {\textgreater{}100} & {\textgreater{}100} & 46.50$\pm$0.50 \\ \bottomrule
\end{tabular}
\end{adjustbox}
\end{table*}

\end{document}